\pdfoutput=1

\documentclass[11pt]{article}

\usepackage[final]{ACL2023}

\usepackage{times}
\usepackage{latexsym}

\usepackage[T1]{fontenc}

\usepackage[utf8]{inputenc}

\usepackage{microtype}

\usepackage{inconsolata}
\usepackage{microtype}
\usepackage{import}
\usepackage{layout}
\usepackage{tabularx, makecell}
\usepackage{booktabs}
\usepackage{mathrsfs}
\usepackage{amssymb} 
\usepackage{url}
\usepackage{hyperref}
\usepackage{graphicx}
\usepackage{xspace,paralist}
\usepackage{times,latexsym}
\usepackage{amsmath}
\usepackage{appendix}
\usepackage{comment} 
\usepackage{enumitem}
\usepackage{makecell}
\usepackage{multirow}
\usepackage[table]{xcolor}
\usepackage{arydshln}
\usepackage{cleveref}
\usepackage{tcolorbox}
\usepackage{todonotes}
\usepackage{longtable,supertabular}
\usepackage{amssymb}
\usepackage{pifont}
\definecolor{bg}{rgb}{0.95,0.95,0.95}
\definecolor{GREEN}{rgb}{0.40,0.60,0.40}

\newcommand{\cmark}{\textcolor{green}{\ding{51}}}
\newcommand{\xmark}{\textcolor{red}{\ding{55}}}

\newcommand{\blue}[1]{\textcolor{blue}{#1}}

\newcommand{\ex}[1]{\textit{#1}\xspace}

\newcommand{\tabref}[1]{Table~\ref{#1}\xspace}
\newcommand{\figref}[1]{Figure~\ref{#1}\xspace}

\newcommand{\name}{\textit{SCALAR}}

\title{SCALAR: Scientific Citation-based Live Assessment \\of Long-context Academic Reasoning}

\author{
Renxi Wang$^{1,2}$\thanks{\hspace{2mm}Equal contributions.} \quad Honglin Mu$^{1,2}$\footnotemark[1] \quad Liqun Ma$^{1}$ \quad Lizhi Lin$^{2,3}$ \quad Yunlong Feng$^{4}$ \\ 
\textbf{Timothy Baldwin$^{1,2,5}$ \quad Xudong Han$^{1,2}$ \quad Haonan Li$^{1,2}$\thanks{\hspace{2mm}Corresponding author.}} \\
$^{1}$MBZUAI \quad $^{2}$LibrAI \quad $^{3}$Tsinghua University \\ $^{4}$Alibaba Group \quad $^{5}$The University of Melbourne \\
}

\begin{document}
\maketitle
\begin{abstract}
Long-context understanding has emerged as a critical capability for large language models (LLMs). However, evaluating this ability remains challenging. We present SCALAR, a benchmark designed to assess citation-grounded long-context reasoning in academic writing. SCALAR leverages academic papers and their citation structure to automatically generate high-quality ground-truth labels without human annotation. It features controllable difficulty levels and a dynamic updating mechanism that mitigates data contamination. 
The benchmark includes two tasks: a multiple-choice QA format and a cloze-style citation prediction. We evaluate a range of state-of-the-art LLMs and find that the multiple-choice task effectively distinguishes model capabilities. While human experts achieve over 90\% accuracy, most models struggle. The cloze-style task is even more challenging, with no model exceeding 50\% accuracy. SCALAR provides a domain-grounded, continuously updating framework for tracking progress in citation-based long-context understanding.\footnote{Code and data are available at \url{https://github.com/LibrAIResearch/scalar}}

\end{abstract}

\section{Introduction}

Large language models (LLMs) have demonstrated impressive capabilities in processing texts of increasing lengths \citep{achiam2023gpt,anthropic2024,dubey2024llama,yang2025qwen2}.
While capable of handling contexts of hundreds of thousands of tokens, evaluating their true understanding of long documents remains challenging. 

Previous evaluations of long-context understanding have often relied on synthetic datasets or simple retrieval tasks like ``needle in a haystack'' variations~\citep{needle,kuratov2024searchneedles11mhaystack,wang2024needle,roberts2024needlethreadingllmsfollow}. While such tasks can test a model's ability to locate information in long sequences, they fail to assess genuine comprehension and are readily solvable by current LLMs \citep{team2024gemini}. Moreover, creating high-quality benchmarks traditionally requires extensive human annotation, which is both time-consuming and costly. Some work transforms short-context tasks into long context by combining them with passages or long documents, such as long-document QA~\citep{needle}, summarization~\citep{chang2024booookscore}, reasoning~\citep{babilong} and reranking~\citep{helmet}. However, such datasets suffer from two key issues: \emph{data contamination} and \emph{shortcut exploitation}, as LLMs can solve problems using their own knowledge rather than the long context. See more related work in \Cref{related_work}.

\begin{figure*}[t]
    \centering
    \tiny
    \includegraphics[width=1.0\linewidth]{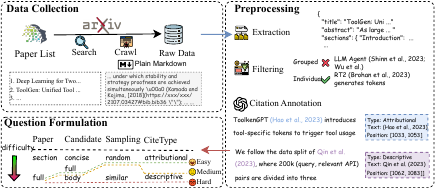}
    \caption{The overall process of building SCALAR. We start with a paper list sourced from arXiv but restrict to papers that have received high scores and been accepted at peer-reviewed conferences. We crawl the raw markdown data (top left). Then we parse them into structured data, sampling citations to mask and other citations as candidates (right). Finally, questions are formulated by masking citations in selected paper sections, choosing candidate spans, applying different sampling strategies, and assigning citation types, aligned with difficulty levels (bottom left).}
    \label{fig:overview}
\end{figure*}

In this work, we present SCALAR, a benchmark designed to evaluate large language models’ (LLMs) citation-grounded long-context reasoning within the scientific domain (\figref{fig:overview}. Our benchmark leverages recently published academic papers and their citations, which are implicitly annotated by domain experts through citation behavior, offering high-quality, unambiguous supervision without manual labeling. By focusing on papers accepted to top-tier venues (e.g., ICLR or ACL) and publicly available through arXiv, SCALAR ensures \textbf{data quality}, \textbf{transparency}, and \textbf{reproducibility}.

To ensure the benchmark remains relevant and minimizes data contamination from model pretraining corpora, we perform a detailed contamination analysis using a 4-year span of ICLR papers (\Cref{sec:data_contamination}). Our findings show that papers prior to 2023 are often memorized by models, while more recent ones (2024–2025) remain largely unseen, underscoring the importance of a live and continuously updated evaluation set. SCALAR is thus designed with a dynamic updating mechanism, allowing automatic integration of the latest high-quality publications.

SCALAR includes two complementary task formats. The first is a multiple-choice citation question answering (MCQA) task, which evaluates a model’s ability to identify the correct cited paper from a set of plausible distractors. The second is a cloze-style citation prediction task, where the model must simultaneously predict four masked citations within a paper by selecting the appropriate references from a shared candidate list. Both tasks are designed to assess citation-grounded reasoning and long-context comprehension across entire scientific documents

To accommodate a wide range of model capabilities, we introduce a flexible difficulty control framework that adjusts both semantic complexity and context length. We define three levels of difficulty (Easy, Medium, and Hard) by systematically varying distractor construction and citation types. This design ensures the benchmark remains suitable for both small and large models.

Extensive experiments across state-of-the-art LLMs reveal substantial performance gaps, especially under longer contexts and harder reasoning settings. To validate task difficulty, we conduct human evaluation on the MCQA task, demonstrating that while humans perform near-perfectly, models still struggle, highlighting critical deficiencies in current long-context modeling.


\begin{table*}[t]
    \centering
    \resizebox{\linewidth}{!}{
    \begin{tabular}{lccccccc}
    \toprule
        \multirow{2}{*}{\textbf{Benchmark}}  & \textbf{Timely} & \multicolumn{2}{c}{\textbf{w/o Potential Issues}} & \textbf{Difficulty} & \textbf{Expert} & \multirow{2}{*}{\textbf{Length}} & \multirow{2}{*}{\textbf{Domain}}\\
        \cmidrule{3-4}
               & \textbf{Update}  & \textbf{Contaminate} & \textbf{Shortcut} & \textbf{Control} & \textbf{Created} & & \\
        \midrule
        Needle-in-haystack \citep{needle} &  \xmark &  \xmark   & \xmark  &  \cmark & \xmark  & Static & N/A\\
        BABILong \citep{babilong}         &  \xmark &  \cmark   & \xmark  &   \cmark & \xmark & Dynamic & Reasoning \\
        RULER \citep{ruler}               &  \xmark &  \cmark   & \xmark  &  \cmark  & \xmark & Dynamic &  General \\
        LongBench \citep{longbench}       &  \xmark &  \xmark   & \xmark  &  \xmark  & \xmark & Dynamic &  General \\
        LongBench-v2 \citep{longbenchv2}  &  \cmark &  \xmark   & \cmark  &  \cmark  & \cmark & Dynamic ($\le$2M) &  General \\
        HELMET \citep{helmet}             &  \xmark &  \cmark   & \cmark  &  \cmark  & \cmark & Dynamic & Multi \\
        SCALAR (ours)             &  \cmark &  \cmark   & \cmark  &  \cmark  & \cmark & Dynamic ($\le$1M) & Academic \\
        \bottomrule
    \end{tabular}}
    \caption{Comparison of long-context benchmarks regarding update frequency, potential data contamination, shortcut susceptibility, difficulty control, expert involvement, length properties, and domain coverage.}
    \label{tab:longcontext-benchmarks}
\end{table*}

\section{Related Work}\label{related_work}

\paragraph{Long-context Evaluation} Many approaches have been proposed to evaluate the ability of language models to utilize a longer context \citep{inftybench,loogle,bamboo,wang2024leavedocumentbehindbenchmarking,song2024countingstarsmultievidencepositionawarescalable}. Real-world evaluations cover long-document QA and summarization \citep{zeroscrolls,laban2024summaryhaystackchallengelongcontext}, mathematics and code understanding \citep{leval,zhao2024docmathevalevaluatingmathreasoning,wang2024mathhayautomatedbenchmarklongcontext}, domain specific analysis \citep{reddy-etal-2024-docfinqa}, and retrieval tasks \citep{helmet}, in various languages \cite{qiu2024clongevalchinesebenchmarkevaluating}, formats \citep{zhang2024marathonracerealmlong}. However, most of these benchmarks repurpose existing corpora, which introduces potential data contamination and limits their ability to measure genuine contextual reasoning.

Meanwhile, benchmarks using synthetic data focus on atomic abilities such as retrieval \citep{needle}, state tracking \cite{babilong}, data aggregation \citep{ruler}, multi-hop reasoning \citep{longbench} like code understanding. Although such datasets provide controllable difficulty, they often diverge from realistic long-context usage.

\textbf{SCALAR complements these efforts} by focusing on \emph{citation-grounded reasoning} within authentic scientific papers. Unlike prior benchmarks, it (i) derives supervision from expert-authored citation links instead of synthetic prompts, (ii) updates dynamically to reduce data contamination, and (iii) provides \emph{explicit difficulty control} through citation type, sampling strategy, and context length. This design bridges the controllability of synthetic datasets with the realism of document-based tasks, enabling a more faithful evaluation of long-context understanding in academic reasoning. For a detailed comparison, see \Cref{tab:longcontext-benchmarks}.

\paragraph{Citation-based Benchmarking} Although scholar literature corpus has been extensively used in language model pretraining \citep{s2orc, pile}, its potential to evaluate long context utilization is not fully explored. A number of datasets focus on generating and recommending citations \citep{citebench, F_rber_2020, 10.1007/978-3-030-99736-6_19}. \citet{litsearch} creates a retrieval benchmark by constructing questions for inline citations using GPT-4 and manually creating questions. There are benchmarks testing models' abilities to answer questions based on papers.  QASPER \citep{qasper} focuses on answering questions about NLP papers, and LitQA \citep{Lala2023PaperQA} examines models' knowledge of biology works.  

These benchmarks capture valuable aspects of academic understanding but are largely confined to local reasoning or single-document contexts. \textbf{SCALAR extends this line of work to the long-context setting} by treating citations as implicit supervision signals for multi-document reasoning. Rather than introducing new annotation pipelines or synthetic prompts, SCALAR systematically repurposes citation structures from recent peer-reviewed papers to assess whether models can link claims in one work to supporting evidence in others. This approach situates citation analysis within a realistic, scalable framework for measuring long-context comprehension.

\section{\name}

This section describes our benchmark construction pipeline. We begin with data collection and preprocessing, including structural parsing and citation filtering. We then detail task formulation and introduce configurable dimensions for difficulty control—covering question scope, citation types, candidate sampling, and candidate representation. We define three difficulty levels, outline the final dataset, and conclude with a data contamination analysis to assess potential training overlap.

\subsection{Data Collection}
We begin by collecting research papers and their corresponding citations from arXiv,\footnote{\url{https://arxiv.org/}} a widely-used open-access repository for scientific papers across various domains.  However, due to its open submission policy, it may include preliminary or non-peer-reviewed low-quality articles. To ensure the quality and reliability of our dataset, we filter for papers that have been accepted by a top-tier venue. More specifically, we select ICLR 2024 and ICLR 2025 accepted papers as original paper list, and search them in arXiv.\footnote{Note that our methods can be generally applied to papers from other venues as well.} We crawl the markdown version as the raw data for further processing.

\subsection{Preprocessing}

\paragraph{Structuralization}
We then extract structured data from raw markdown texts. This includes identifying the title, abstract, and sections of the paper, cleaning links, and extract citations. Each citation is marked with its position and mapped to its corresponding reference in the bibliography. After that, we use the same data collection process to gather information about the cited papers. We separate papers into top-level sections, finding that papers in our dataset contain an average of 6.1 sections. For all papers, we remove references and appendices, leaving out the main content for task formulation.

\paragraph{Citation Filtering} To ensure high-quality evaluation data, we distinguish between two categories of citations: grouped citations and individual citations. \textbf{Grouped Citations} refer to multiple sources together in a single parenthetical reference, typically used when summarizing general information, such as ``\textit{prior works (Liu et al., 2024; Hsieh et al., 2024a; Zhang et al., 2024)}''. \textbf{Individual Citations}, in contrast, are used to refer to single source separately, usually when discussing specific methods or results, for instance ``\textit{needle in a haystack' test (Kamradt, 2023)}''. Since grouped citations may have multiple correct answers when masked one of them, during the processing in \Cref{sec:task_formulation}, we exclusively sample and mask individual citations to maintain clear ground truth labels.


\subsection{Task Formulation}
\label{sec:task_formulation}

We define two citation resolution tasks designed to evaluate a language model’s ability to identify masked references in scientific papers: \textbf{Multiple-choice QA} (MCQA) and \textbf{Cloze Test}. Both tasks involve replacing in-text citations with placeholders and require the model to recover the correct references using the surrounding context. While Multiple-choice QA focuses on identifying a single masked citation from a set of options, the Cloze Test requires resolving multiple masked citations jointly.

For both tasks, each test case comprises three distinct parts:
\begin{itemize}
    \item \textbf{Instruction} provides task-specific guidance, including how the LLM should complete the task, the expected answer format, and the roles of other components. 
    \item \textbf{Question Paper} contains either the full text or a specific section of a paper, with one or more citations replaced by placeholders such as \texttt{**[MASKED\_CITATION]**}.
    \item \textbf{Candidates} list four reference options. In the case of Multiple-choice QA, this includes one correct answer and three distractors.
\end{itemize}
To ensure high annotation quality, we draw all candidate references from the reference list of the question paper. This design leverages the expertise of the original authors, who have carefully selected relevant prior work for citation. By restricting candidates to this curated list, we avoid introducing external papers that, while potentially topically similar, may not align with the author’s intent and could be more suitable for the citation than the actual ground truth.

In our prompt implementation, we define several XML elements to separate different elements. The details of prompt templates are shown in \Cref{sec:prompt}.

\begin{figure}[t]
\centering
    \includegraphics[width=0.9\linewidth]{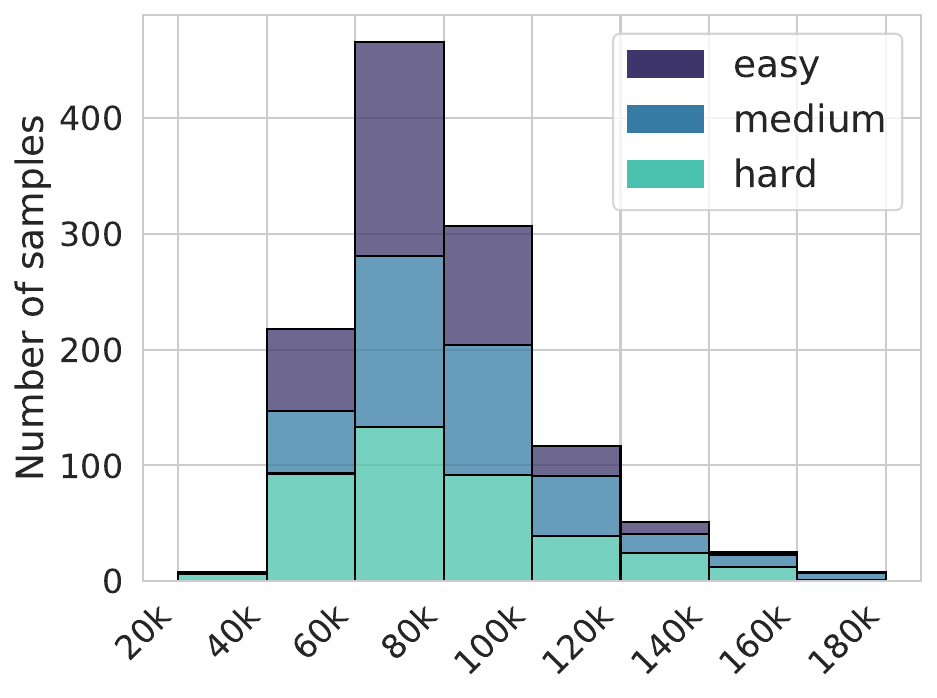}
    \caption{Token length distribution of the three subsets of our dataset. Samples are tokenized using GPT-4o tokenizer from tiktoken \citep{openai2023tiktoken}.}
	\label{fig:length_dist}
\end{figure}

\subsection{Difficulty Configuration Dimensions}
\label{sec:difficulty-dimensions}

We define four core configuration dimensions that govern each benchmark instance: question paper scope, citation type, distractor sampling strategy, and candidate representation. These dimensions allow fine-grained control over both semantic complexity and context length. A visual summary is shown in \Cref{sec:difficulty_config}.

\paragraph{Question Paper Scope.} We use two modes for presenting the source paper to the model. In the Single Section setting, we provide only the section that contains the masked citation, thereby limiting contextual cues. In the Full Paper setting, the entire paper is included as context, offering broader information for reasoning.

\paragraph{Citation Type.} We distinguish between two types of citations based on how they are presented in the text. \textbf{Attributional Citations} explicitly mention a specific model, method, or dataset (e.g., \ex{BERT (Devlin et al., [2018]) is used for embedding texts...}). In contrast, \textbf{Descriptive Citations} embed the reference in a more narrative form without explicitly naming it (e.g., \ex{Pretraining a bidirectional transformer (Devlin et al., [2018]) is time-consuming...}).

\paragraph{Candidate Sampling.} To construct challenging MCQA and cloze test options, we consider two candidate sampling strategies. In the Random Sampling setting, candidates are randomly drawn from the reference list of the question paper. In the Nearest Sampling setting, we sample four references that are cited at the same section. For MCQA, we then randomly mask one citation corresponding to one reference, and take other references as distractors. For cloze test, we mask four citations in the question paper, each corresponding to one of the references.

\paragraph{Candidate Representation.} Each candidate reference can be presented in one of three formats. The Concise format includes only the title and abstract. The Full format includes the complete content of the cited paper. The Body format removes the title, abstract, introduction, and conclusion, leaving only the main body text to test the model’s reasoning under limited cues.

\subsection{Difficulty Levels and Final Dataset}
\label{sec:difficuly}

Building on these dimensions, we define three standard difficulty levels to facilitate controlled benchmarking. These levels progressively increase in complexity by varying citation types, candidate selection strategies, and input lengths:  (I) \textbf{Easy} level samples candidates randomly, and only mask attributional citations. (II) \textbf{Medium} level also samples candidates randomly while only masks descriptive citations. Both easy and medium level use the full paper for the question paper and candidates. (III) \textbf{Hard} level masks descriptive citations, but candidates are sampled from nearest references. Additionally, we use only the body of the paper for candidates, to avoid the model easily get answer from titles, abstracts, introductions, and conclusions.

\begin{figure}[t]
\centering
    \includegraphics[width=1.01\linewidth]{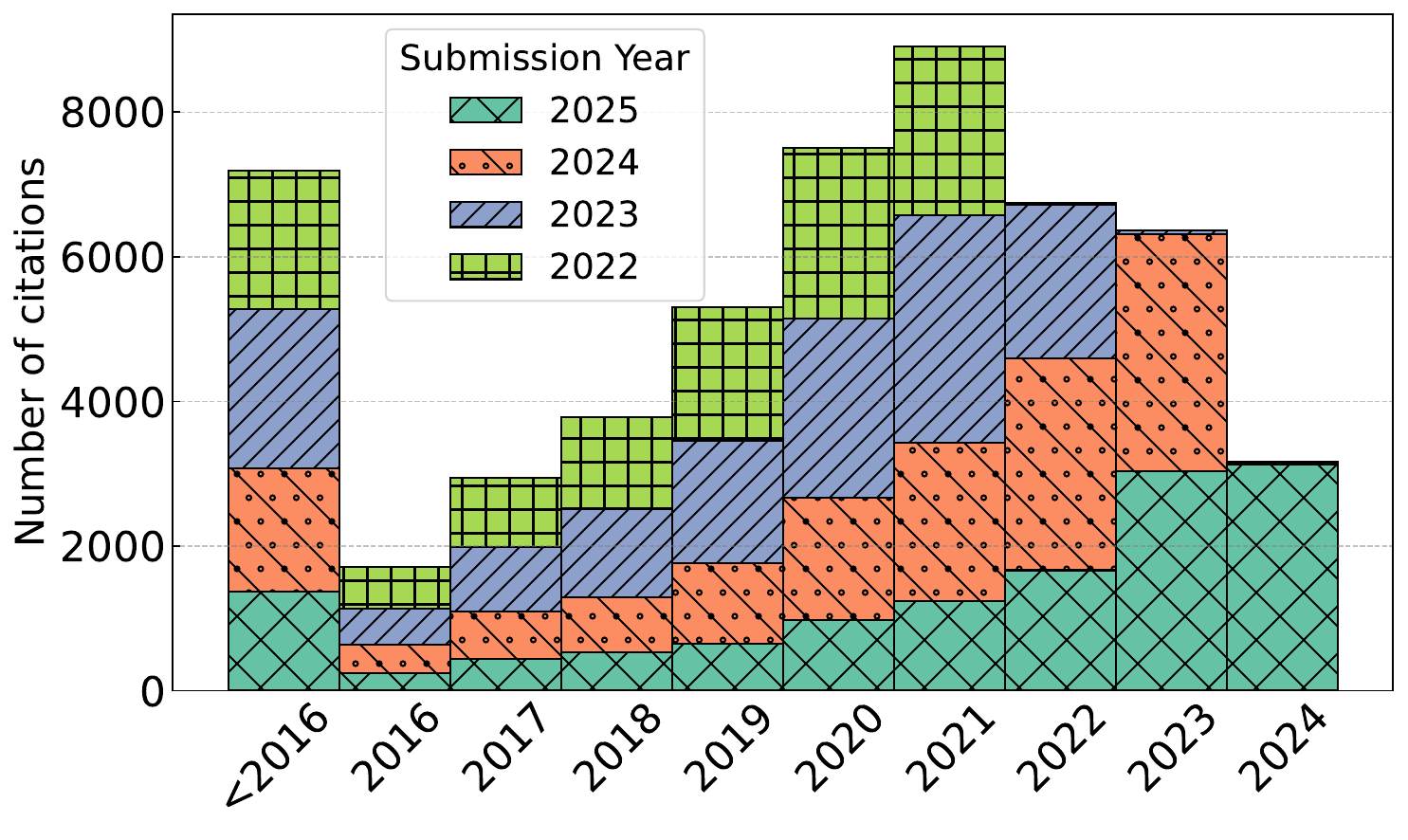}
    \caption{Distribution of cited papers' publication years across ICLR submission years used in SCALAR.}
	\label{fig:year_dist}
\end{figure}

The final dataset (per year) consists of 600 questions evenly distributed across two tasks and three difficulty levels, with each question containing four candidates. To ensure diversity, we limit at most five questions per paper. All papers, including question paper and candidates, are limited to 100,000 characters to accommodate model context limitations, with questions formatted using the template in \Cref{sec:prompt}. 

We collect data from papers published between 2022 and 2025; while data from 2022 and 2023 are primarily used for retrospective analysis, our benchmark focuses on 2024 and 2025 to emphasize its live and forward-looking nature. The length distribution is shown in \Cref{fig:length_dist}. Most samples have tokens between 60k and 80k. Hard set are slightly shorter than Easy and Medium set, since we only use the body content of the candidate papers. Figure~\ref{fig:year_dist} shows the submission or publication years of cited papers for specific venues. As shown in figure, most papers tend to cite more recent papers.

\begin{figure}[t]    
\centering
    \includegraphics[width=1.01\linewidth]{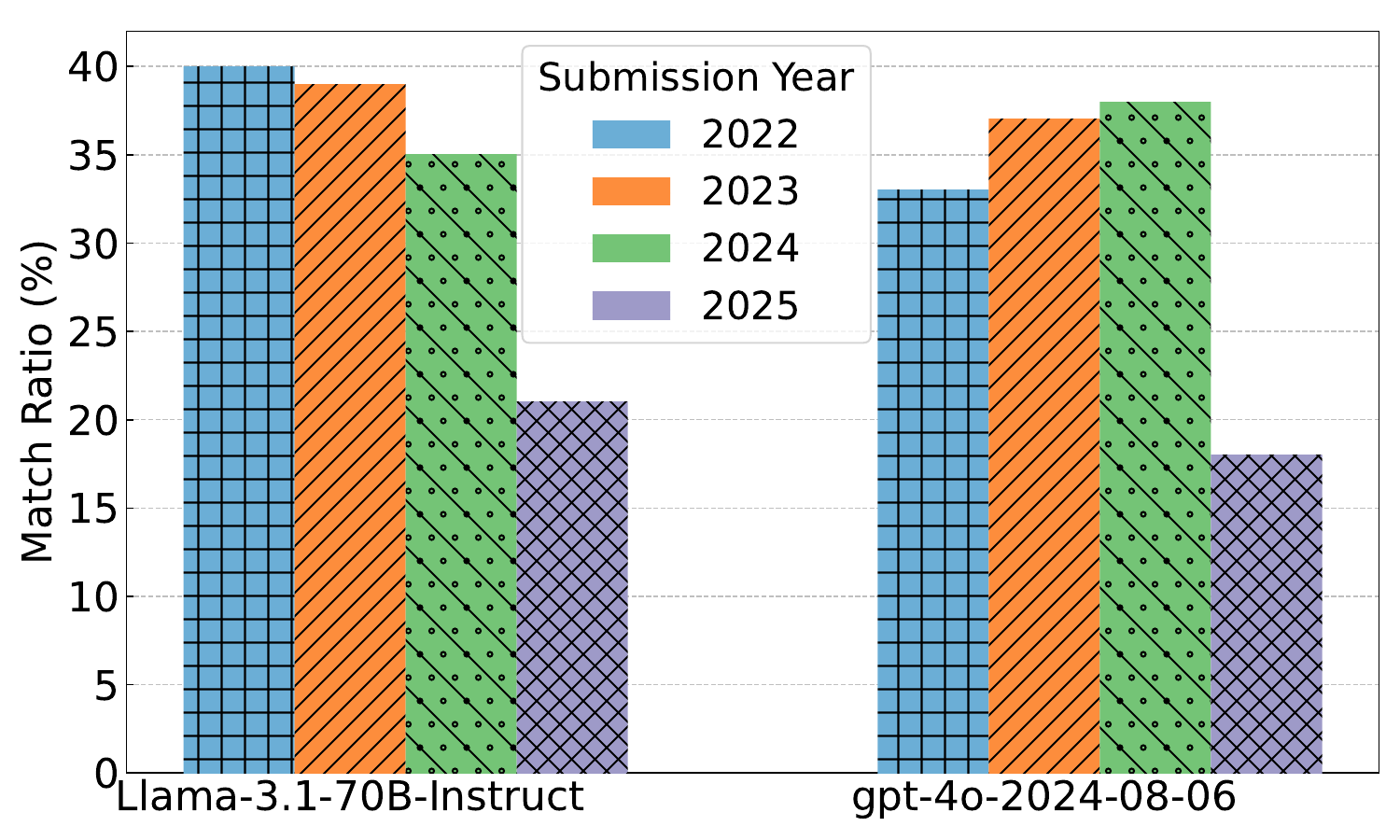}
    \caption{Data contamination prediction across submission years. We report the match rate (\%), defined as the proportion of ground-truth citations successfully predicted by the model when prompted with the citation context. Higher match rates suggest a greater likelihood of data contamination.}
	\label{fig:data_contamination}
\end{figure}

\definecolor{lightblue}{rgb}{0.85,0.92,1}
\definecolor{green}{rgb}{0.5, 1, 0.5}
\definecolor{red}{rgb}{1,0.5,0.5}

\begin{table*}[t]
\small
\centering
\resizebox{0.9\linewidth}{!}{
\begin{tabular}{lccccccc}
\toprule
\multirow{2}{*}{Model} & \multicolumn{3}{c}{ICLR 2024} & \multicolumn{3}{c}{ICLR 2025} &  \multirow{2}{*}{Average} \\
\cmidrule{2-4}\cmidrule{5-7}
& Easy & Medium & Hard & Easy & Medium & Hard & \\
\midrule
Llama-3.1-8B & \cellcolor{green!21!red!20!white}0.21 & \cellcolor{green!22!red!20!white}0.22 & \cellcolor{green!23!red!20!white}0.23 & \cellcolor{green!30!red!20!white}0.30 & \cellcolor{green!28!red!20!white}0.28 & \cellcolor{green!24!red!20!white}0.24 & \cellcolor{green!25!red!20!white}0.25 \\
Llama-3.1-70B & \cellcolor{green!37!red!20!white}0.37 & \cellcolor{green!25!red!20!white}0.25 & \cellcolor{green!30!red!20!white}0.30 & \cellcolor{green!37!red!20!white}0.37 & \cellcolor{green!29!red!20!white}0.29 & \cellcolor{green!16!red!20!white}0.16 & \cellcolor{green!29!red!20!white}0.29 \\
Llama-3.3-70B & \cellcolor{green!43!red!20!white}0.43 & \cellcolor{green!28!red!20!white}0.28 & \cellcolor{green!29!red!20!white}0.29 & \cellcolor{green!37!red!20!white}0.37 & \cellcolor{green!36!red!20!white}0.36 & \cellcolor{green!23!red!20!white}0.23 & \cellcolor{green!33!red!20!white}0.33 \\
Qwen2.5-7B-1M & \cellcolor{green!31!red!20!white}0.31 & \cellcolor{green!34!red!20!white}0.34 & \cellcolor{green!22!red!20!white}0.22 & \cellcolor{green!49!red!20!white}0.49 & \cellcolor{green!38!red!20!white}0.38 & \cellcolor{green!28!red!20!white}0.28 & \cellcolor{green!34!red!20!white}0.34 \\
Phi-3-Mini & \cellcolor{green!19!red!20!white}0.19 & \cellcolor{green!21!red!20!white}0.21 & \cellcolor{green!26!red!20!white}0.26 & \cellcolor{green!39!red!20!white}0.39 & \cellcolor{green!21!red!20!white}0.21 & \cellcolor{green!20!red!20!white}0.20 & \cellcolor{green!24!red!20!white}0.24 \\
Phi-3-Medium & \cellcolor{green!12!red!20!white}0.12 & \cellcolor{green!17!red!20!white}0.17 & \cellcolor{green!12!red!20!white}0.12 & \cellcolor{green!20!red!20!white}0.20 & \cellcolor{green!12!red!20!white}0.12 & \cellcolor{green!8!red!20!white}0.08 & \cellcolor{green!14!red!20!white}0.14 \\
Claude-3.5-Haiku & \cellcolor{green!71!red!20!white}0.71 & \cellcolor{green!56!red!20!white}0.56 & \cellcolor{green!44!red!20!white}0.44 & \cellcolor{green!77!red!20!white}0.77 & \cellcolor{green!61!red!20!white}0.61 & \cellcolor{green!42!red!20!white}0.42 & \cellcolor{green!58!red!20!white}0.58 \\
GPT-4o-Mini & \cellcolor{green!75!red!20!white}0.75 & \cellcolor{green!51!red!20!white}0.51 & \cellcolor{green!38!red!20!white}0.38 & \cellcolor{green!81!red!20!white}0.81 & \cellcolor{green!56!red!20!white}0.56 & \cellcolor{green!48!red!20!white}0.48 & \cellcolor{green!58!red!20!white}0.58 \\
GPT-4o & \cellcolor{green!92!red!20!white}0.92 & \cellcolor{green!66!red!20!white}0.66 & \cellcolor{green!56!red!20!white}0.56 & \cellcolor{green!95!red!20!white}0.95 & \cellcolor{green!72!red!20!white}0.72 & \cellcolor{green!50!red!20!white}0.50 & \cellcolor{green!72!red!20!white}0.72 \\
\midrule
QwQ-32B & \cellcolor{green!50!red!20!white}0.50 & \cellcolor{green!42!red!20!white}0.42 & \cellcolor{green!45!red!20!white}0.45 & \cellcolor{green!47!red!20!white}0.47 & \cellcolor{green!46!red!20!white}0.46 & \cellcolor{green!46!red!20!white}0.46 & \cellcolor{green!46!red!20!white}0.46 \\
DeepSeek-R1-0528 &  \cellcolor{green!93!red!20!white}0.93 & \cellcolor{green!78!red!20!white}0.78 & \cellcolor{green!71!red!20!white}0.71 & \cellcolor{green!96!red!20!white}0.96 & \cellcolor{green!89!red!20!white}0.89 & \cellcolor{green!66!red!20!white}0.66 & \cellcolor{green!82!red!20!white}0.82 \\
Kimi-k2-Thinking &  \cellcolor{green!95!red!20!white}0.95 & \cellcolor{green!87!red!20!white}0.87 & \cellcolor{green!82!red!20!white}0.82 & \cellcolor{green!99!red!20!white}0.99 & \cellcolor{green!91!red!20!white}0.91 & \cellcolor{green!76!red!20!white}0.76 & \cellcolor{green!88!red!20!white}0.88 \\
Gemini-2.5-Flash &  \cellcolor{green!95!red!20!white}0.95 & \cellcolor{green!73!red!20!white}0.73 & \cellcolor{green!74!red!20!white}0.74 & \cellcolor{green!94!red!20!white}0.94 & \cellcolor{green!77!red!20!white}0.77 & \cellcolor{green!61!red!20!white}0.61 & \cellcolor{green!79!red!20!white}0.79 \\
Gemini-2.5-Pro &  \cellcolor{green!98!red!20!white}0.98 & \cellcolor{green!92!red!20!white}0.92 & \cellcolor{green!93!red!20!white}0.93 & \cellcolor{green!99!red!20!white}0.99 & \cellcolor{green!92!red!20!white}0.92 & \cellcolor{green!85!red!20!white}0.85 & \cellcolor{green!93!red!20!white}0.93 \\
GPT-5-mini-medium &  \cellcolor{green!99!red!20!white}0.99 & \cellcolor{green!85!red!20!white}0.85 & \cellcolor{green!90!red!20!white}0.90 & \cellcolor{green!99!red!20!white}0.99 & \cellcolor{green!92!red!20!white}0.92 & \cellcolor{green!82!red!20!white}0.82 & \cellcolor{green!91!red!20!white}0.91 \\
GPT-5-mini-high &  \cellcolor{green!99!red!20!white}0.99 & \cellcolor{green!87!red!20!white}0.87 & \cellcolor{green!90!red!20!white}0.90 & \cellcolor{green!99!red!20!white}0.99 & \cellcolor{green!92!red!20!white}0.92 & \cellcolor{green!81!red!20!white}0.81 & \cellcolor{green!91!red!20!white}0.91 \\
GPT-5-high &  \cellcolor{green!99!red!20!white}0.99 & \cellcolor{green!95!red!20!white}0.95 & \cellcolor{green!94!red!20!white}0.94 & \cellcolor{green!99!red!20!white}0.99 & \cellcolor{green!95!red!20!white}0.95 & \cellcolor{green!92!red!20!white}0.92 & \cellcolor{green!96!red!20!white}0.96 \\
o4-mini-medium &  \cellcolor{green!98!red!20!white}0.98 & \cellcolor{green!88!red!20!white}0.88 & \cellcolor{green!81!red!20!white}0.81 & \cellcolor{green!99!red!20!white}0.99 & \cellcolor{green!88!red!20!white}0.88 & \cellcolor{green!72!red!20!white}0.72 & \cellcolor{green!88!red!20!white}0.88 \\
o4-mini-high &  \cellcolor{green!97!red!20!white}0.97 & \cellcolor{green!87!red!20!white}0.87 & \cellcolor{green!86!red!20!white}0.86 & \cellcolor{green!99!red!20!white}0.99 & \cellcolor{green!88!red!20!white}0.88 & \cellcolor{green!74!red!20!white}0.74 & \cellcolor{green!89!red!20!white}0.89 \\
\bottomrule
\end{tabular}
}
\caption{Multiple-choice question answering accuracy on ICLR 2024 and ICLR 2025 datasets, evaluated across three difficulty levels. Cell colors indicate performance from \textcolor{red}{low (red)} to \textcolor{GREEN}{high (green)}.}
\label{tab:mcqa_results}
\end{table*}

\section{Data Contamination Analysis}\label{sec:data_contamination}
To evaluate potential data contamination, we conduct a targeted analysis using a subset of hard-level questions drawn from our benchmark. For each question, we input the full context from the question paper up to, but not including, the masked citation. We then prompt two language models to generate 20 tokens and assess whether the ground truth author name is explicitly predicted within the generated sequence.

The models we evaluate include 
\textit{LLaMA-3.1-70B-Instruct} (training cutoff: December 2023) and \textit{GPT-4o} (training cutoff: June 2024). We record the proportion of questions where the correct author name is successfully predicted by each model.

Figure~\ref{fig:data_contamination} illustrates the results across publication years. All models demonstrate a high success rate in predicting citations from papers in earlier years, with markedly reduced accuracy for papers published in 2025. This temporal trend strongly suggests that earlier papers are more likely to have been part of the models’ training corpora.
These findings highlight the persistent risk of data contamination in static benchmarks and reinforce the importance of using a live, dynamically updated benchmark to ensure fair and forward-looking evaluation.

\section{Experiments}

We benchmark a set of open-source and proprietary models with long-context capabilities ($\geq 128k$ tokens) on SCALAR.

\paragraph{Open-source models}
(1) \textit{Llama-3.1-8B-Instruct}, \textit{Llama-3.1-70B-Instruct}, and \textit{Llama-3.3-70B-Instruct} are variants from Meta’s Llama series~\citep{touvron2023llama2, dubey2024llama}, each supporting 128K-token contexts.
(2) \textit{Qwen2.5-7B-1M} \citep{yang2025qwen2} is a model from Alibaba with long context support.
(3) \textit{Phi-3-Mini-128K-Instruct} and \textit{Phi-3-Medium-128K-Instruct} \citep{abdin2024phi4technicalreport} are lightweight models from Microsoft designed for efficiency, both supporting 128K tokens.

\paragraph{Proprietary models}
(4) \textit{Claude-3.5-Haiku} \citep{claude21modelcard} is Anthropic’s high-speed long-context model with a 200K-token context window.
(5)  \textit{GPT-4o-Mini} and \textit{GPT-4o} \citep{achiam2023gpt} are proprietary models from OpenAI, each with 128K-token context limits. GPT-4o represents OpenAI’s most advanced model to date.

Recently, a new class of reasoning-focused models has emerged, such as DeepSeek-R1 \citep{guo2025deepseek} and others \citep{yu2025dapoopensourcellmreinforcement,openr1}, which emphasize reasoning capabilities through reinforcement learning. These models are designed specifically to enhance logical inference, multi-hop question answering, and mathematical reasoning. Although some of them were released within three months of our benchmark creation, we also include their performance for reference.



\begin{table*}[t]
\small
\centering
\resizebox{0.9\linewidth}{!}{
\begin{tabular}{lccccccc}
\toprule
\multirow{2}{*}{Model} & \multicolumn{3}{c}{ICLR 2024} & \multicolumn{3}{c}{ICLR 2025} &  \multirow{2}{*}{Average} \\
\cmidrule{2-4}\cmidrule{5-7}
& Easy & Medium & Hard & Easy & Medium & Hard & \\
\midrule
Llama-3.1-8B & \cellcolor{green!24!red!20!white}0.24 & \cellcolor{green!24!red!20!white}0.24 & \cellcolor{green!25!red!20!white}0.25 & \cellcolor{green!24!red!20!white}0.24 & \cellcolor{green!20!red!20!white}0.20 & \cellcolor{green!20!red!20!white}0.20 & \cellcolor{green!23!red!20!white}0.23 \\
Llama-3.1-70B & \cellcolor{green!28!red!20!white}0.28 & \cellcolor{green!19!red!20!white}0.19 & \cellcolor{green!21!red!20!white}0.21 & \cellcolor{green!23!red!20!white}0.23 & \cellcolor{green!23!red!20!white}0.23 & \cellcolor{green!19!red!20!white}0.19 & \cellcolor{green!22!red!20!white}0.22 \\
Llama-3.3-70B & \cellcolor{green!27!red!20!white}0.27 & \cellcolor{green!18!red!20!white}0.18 & \cellcolor{green!22!red!20!white}0.22 & \cellcolor{green!22!red!20!white}0.22 & \cellcolor{green!22!red!20!white}0.22 & \cellcolor{green!18!red!20!white}0.18 & \cellcolor{green!22!red!20!white}0.22 \\
Qwen2.5-7B-1M & \cellcolor{green!30!red!20!white}0.30 & \cellcolor{green!21!red!20!white}0.21 & \cellcolor{green!24!red!20!white}0.24 & \cellcolor{green!25!red!20!white}0.25 & \cellcolor{green!28!red!20!white}0.28 & \cellcolor{green!24!red!20!white}0.24 & \cellcolor{green!25!red!20!white}0.25 \\
Phi-3-Mini & \cellcolor{green!20!red!20!white}0.20 & \cellcolor{green!19!red!20!white}0.19 & \cellcolor{green!16!red!20!white}0.16 & \cellcolor{green!18!red!20!white}0.18 & \cellcolor{green!16!red!20!white}0.16 & \cellcolor{green!21!red!20!white}0.21 & \cellcolor{green!18!red!20!white}0.18 \\
Phi-3-Medium & \cellcolor{green!14!red!20!white}0.14 & \cellcolor{green!11!red!20!white}0.11 & \cellcolor{green!12!red!20!white}0.12 & \cellcolor{green!12!red!20!white}0.12 & \cellcolor{green!12!red!20!white}0.12 & \cellcolor{green!13!red!20!white}0.13 & \cellcolor{green!12!red!20!white}0.12 \\
Claude-3.5-Haiku & \cellcolor{green!37!red!20!white}0.37 & \cellcolor{green!26!red!20!white}0.26 & \cellcolor{green!26!red!20!white}0.26 & \cellcolor{green!31!red!20!white}0.31 & \cellcolor{green!30!red!20!white}0.30 & \cellcolor{green!31!red!20!white}0.31 & \cellcolor{green!30!red!20!white}0.30 \\
GPT-4o-Mini & \cellcolor{green!35!red!20!white}0.35 & \cellcolor{green!22!red!20!white}0.22 & \cellcolor{green!28!red!20!white}0.28 & \cellcolor{green!35!red!20!white}0.35 & \cellcolor{green!28!red!20!white}0.28 & \cellcolor{green!28!red!20!white}0.28 & \cellcolor{green!29!red!20!white}0.29 \\
GPT-4o & \cellcolor{green!45!red!20!white}0.45 & \cellcolor{green!30!red!20!white}0.30 & \cellcolor{green!35!red!20!white}0.35 & \cellcolor{green!42!red!20!white}0.42 & \cellcolor{green!36!red!20!white}0.36 & \cellcolor{green!32!red!20!white}0.32 & \cellcolor{green!37!red!20!white}0.37 \\
\midrule
QwQ-32B & \cellcolor{green!40!red!20!white}0.40 & \cellcolor{green!29!red!20!white}0.29 & \cellcolor{green!34!red!20!white}0.34 & \cellcolor{green!34!red!20!white}0.34 & \cellcolor{green!29!red!20!white}0.29 & \cellcolor{green!34!red!20!white}0.34 & \cellcolor{green!33!red!20!white}0.33 \\
DeepSeek-R1-0528 &  \cellcolor{green!35!red!20!white}0.35 & \cellcolor{green!21!red!20!white}0.21 & \cellcolor{green!25!red!20!white}0.25 & \cellcolor{green!28!red!20!white}0.28 & \cellcolor{green!25!red!20!white}0.25 & \cellcolor{green!22!red!20!white}0.22 & \cellcolor{green!26!red!20!white}0.26 \\
Gemini-2.5-Flash &  \cellcolor{green!42!red!20!white}0.42 & \cellcolor{green!37!red!20!white}0.37 & \cellcolor{green!38!red!20!white}0.38 & \cellcolor{green!35!red!20!white}0.35 & \cellcolor{green!44!red!20!white}0.44 & \cellcolor{green!40!red!20!white}0.40 & \cellcolor{green!39!red!20!white}0.39 \\
Gemini-2.5-Pro &  \cellcolor{green!44!red!20!white}0.44 & \cellcolor{green!43!red!20!white}0.43 & \cellcolor{green!39!red!20!white}0.39 & \cellcolor{green!35!red!20!white}0.35 & \cellcolor{green!49!red!20!white}0.49 & \cellcolor{green!48!red!20!white}0.48 & \cellcolor{green!43!red!20!white}0.43 \\
GPT-5-mini-medium &  \cellcolor{green!43!red!20!white}0.43 & \cellcolor{green!34!red!20!white}0.34 & \cellcolor{green!32!red!20!white}0.32 & \cellcolor{green!34!red!20!white}0.34 & \cellcolor{green!40!red!20!white}0.40 & \cellcolor{green!35!red!20!white}0.35 & \cellcolor{green!36!red!20!white}0.36 \\
GPT-5-mini-high &  \cellcolor{green!43!red!20!white}0.43 & \cellcolor{green!35!red!20!white}0.35 & \cellcolor{green!32!red!20!white}0.32 & \cellcolor{green!35!red!20!white}0.35 & \cellcolor{green!41!red!20!white}0.41 & \cellcolor{green!37!red!20!white}0.37 & \cellcolor{green!37!red!20!white}0.37 \\
GPT-5-high &  \cellcolor{green!44!red!20!white}0.44 & \cellcolor{green!34!red!20!white}0.34 & \cellcolor{green!38!red!20!white}0.38 & \cellcolor{green!35!red!20!white}0.35 & \cellcolor{green!43!red!20!white}0.43 & \cellcolor{green!42!red!20!white}0.42 & \cellcolor{green!39!red!20!white}0.39 \\
o4-mini-medium &  \cellcolor{green!43!red!20!white}0.43 & \cellcolor{green!35!red!20!white}0.35 & \cellcolor{green!26!red!20!white}0.26 & \cellcolor{green!34!red!20!white}0.34 & \cellcolor{green!45!red!20!white}0.45 & \cellcolor{green!36!red!20!white}0.36 & \cellcolor{green!37!red!20!white}0.37 \\
o4-mini-high &  \cellcolor{green!42!red!20!white}0.42 & \cellcolor{green!36!red!20!white}0.36 & \cellcolor{green!31!red!20!white}0.31 & \cellcolor{green!34!red!20!white}0.34 & \cellcolor{green!40!red!20!white}0.40 & \cellcolor{green!36!red!20!white}0.36 & \cellcolor{green!37!red!20!white}0.37 \\
\bottomrule
\end{tabular}
}
\caption{Cloze-style citation prediction scores for non-reasoning and reasoning LLMs across two venues (ICLR 2024 and 2025). Cell colors indicate performance from \textcolor{red}{low (red)} to \textcolor{GREEN}{high (green)}.}
\label{tab:cloze_results}
\end{table*}

\subsection{Main Results}

\paragraph{Multiple-choice QA}
\Cref{tab:mcqa_results} presents model performance across difficulty levels.  Generally, all LLMs' performance downgrades when difficulty increases. Reasoning LLMs and non-reasoning LLMs show a large gap in performance, indicating that our benchmark requires reasoning capability to finish. 

For non-reasoning LLMs, on the easy level, \name\ can already differentiate their long context capability, where the best model GPT-4o achieves 95\% accuracy, while most models get a score lower than 60\%, compared to the random baseline of 25\%. For the hard level set, half of the models achieve random performance, and even current SOTA models obtain less than half the correct results, demonstrating how challenging our dataset is. 
The below-random performance of the Phi models is largely attributed to their failure to adhere to the expected output format, which leads to unsuccessful answer extraction. The hard level proves particularly challenging. Even state-of-the-art models like GPT-4o achieve only 50\% accuracy, while most other models perform near random chance. 

For reasoning LLMs, most of them achieved better performance than non-reasoning models. And the performance drop as the difficulty level increases is not as high as non-reasoning LLMs. The best model, GPT-5 with high reasoning effort achieved over 90\% accuracy on all subsets of our MCQ benchmark.


\paragraph{Cloze-style QA}
As depicted in \Cref{tab:cloze_results}, the cloze-style test setting presents a greater difficulty to language models compared to multiple-choice formulations, primarily because they require the prediction of four citations at once. This increased challenge is evident in performance shifts: the accuracy of most open-source models degrades by approximately 10\% relative to their multiple-choice QA performance, while proprietary models can see their scores reduced by as much as half. Overall, even the best-performing model achieves an average accuracy of less than 40\%.

\begin{table}[t]
    \centering
    \begin{tabular}{cccc}
    \toprule
        Annotator & Easy & Medium & Hard \\
        \midrule
        Human 1 & 1.00 & 0.90 & 0.90 \\
        Human 2 & 1.00 & 0.80 & 0.80 \\ 
        \bottomrule
    \end{tabular}
    \caption{Performance of human experts on the multiple-choice QA task. A total of 30 questions were evaluated, evenly drawn across three difficulty levels. Human experts achieved perfect or near-perfect accuracy.}
    \label{tab:human_eval}
\end{table}

\paragraph{Human Performance on MCQA}\label{sec:human}
To better understand the task difficulty from a human perspective, we evaluated expert performance on the ICLR 2025 portion of our benchmark. Specifically, two PhD candidates in computer science were asked to complete a multiple-choice QA task consisting of 30 questions evenly sampled across the three predefined difficulty levels. As shown in \Cref{tab:human_eval}, both annotators achieved perfect accuracy on the easy subset and high accuracy on the medium and hard subsets. These results demonstrate that while the task challenges current LLMs, it remains straightforward for human experts familiar with scientific content.

\begin{table}[t]
    \centering
    \resizebox{\linewidth}{!}{
    \begin{tabular}{cccccc}
    \toprule
        \multirow{2}{*}{Question}  & \multirow{2}{*}{Candidate} & \multicolumn{2}{c}{Easy} & \multicolumn{2}{c}{Hard} \\
        \cmidrule{3-4} \cmidrule{5-6}
        & & GPT & Qwen & GPT & Qwen \\
        \midrule
        Section& Concise & 0.86 & 0.88 & 0.58 & 0.48\\
        Full   & Concise & 0.86 & 0.72 & 0.52 & 0.40\\
        Section& Full    & 0.80 & 0.40 & 0.64 & 0.22\\
        Full   & Full    & 0.80 & 0.42 & 0.50 & 0.24\\
        \bottomrule
    \end{tabular}}
    \caption{Impact of context length on model performance. Results compare GPT-4o-mini and Qwen2.5-7B-Instruct-1M across different configurations. \textbf{Easy} level uses attributional citations with random candidates, while \textbf{Hard} level uses descriptive citations with nearest-neighbor candidates.}
    \label{tab:length}
\end{table}

\begin{table}[t]
    \centering
    \resizebox{\linewidth}{!}{
    \begin{tabular}{llcccc}
    \toprule
        Cite Type& Sampling &Candidate & GPT & Qwen  \\
        \midrule
        Attributional &Random & Full &  0.82 & 0.42\\
        Attributional &Nearest& Full &  0.82 & 0.38\\
        Descriptive   &Random & Full &  0.56 & 0.34\\
        Descriptive   &Nearest& Full &  0.52 & 0.22\\
        Descriptive   &Nearest& Body &  0.44 & 0.28\\
        \bottomrule
    \end{tabular}}
    \caption{Impact of semantic reasoning difficulty on model performance. Results of GPT-4o-mini and Qwen2.5-7B-Instruct-1M.}
    \label{tab:semantic}
\end{table}

\subsection{Error Analysis}
We manually examined 72 incorrect predictions across models and difficulty levels. The analysis reveals several consistent failure patterns: (1) Overreliance on semantic similarity. In over 70\% of the incorrect cases, models selected papers sharing the same research field or similar methods, even when they were irrelevant to the actual citation. (2) Failure in citation-level reasoning: some cases misinterpret content near the citation location---over half of the incorrect predictions had irrelevant or weakly related citation context. (3) There are some model-specific trends. For example,  LLaMA series frequently confused weakly related papers, even when the correct answer had distinct content. Also showed occasional format-following issues. GPT-4o, Claude, and Qwen are relatively more robust, but still failed when distractors shared close topical similarity or when keywords near the citation misled the model.

These findings suggest that while models can capture global topical alignment, they often struggle with fine-grained, context-sensitive citation understanding—underscoring the value of SCALAR in evaluating deep reasoning over long contexts.

\subsection{Difficulty Analysis}
Following our discussion of difficulty control in \Cref{sec:difficuly}, where we established three difficulty levels, we now conduct a detailed analysis of how different configuration combinations affect task difficulty. We categorize these configurations into two primary dimensions: context length and semantic complexity. For each configuration setting in our analysis, we evaluate model performance on a sample of 50 questions.

\paragraph{Context Length}
We analyze the impact of context length by varying both question and candidate paper representations, as shown in \Cref{tab:length}. For questions, we compare using either the full paper or just the section containing the masked citation. For candidates, we compare using either the full paper or just the abstract. This creates four distinct length configurations while controlling for semantic difficulty.

The results show that both models generally perform better with more concise contexts. GPT-4o-mini maintains relatively stable performance across configurations, while Qwen2.5-7B shows significant degradation when given full candidate papers (dropping from 88\% to 40-42\% in the easy setting). This pattern persists in the hard setting, though with overall lower performance, suggesting that focused, relevant context may be more beneficial than comprehensive but potentially noisy full-paper information.

\paragraph{Semantic Complexity}
\Cref{tab:semantic} demonstrates how different semantic factors affect model performance on citation prediction. We analyze three key dimensions: citation type (attributional vs. descriptive), candidate sampling method (random vs. nearest), and candidate representation (full paper vs. body only).

The results show a clear hierarchy of difficulty. Both models perform best with attributional citations, likely due to their more straightforward nature. Performance drops significantly with descriptive citations, particularly when combined with nearest-neighbor sampling.
While GPT-4o-mini maintains above-random performance across all configurations, Qwen's performance on the most challenging setting (descriptive, nearest sampling, full paper) drops to 22\%, below random chance (25\%). Interestingly, when using body-only candidates, Qwen's performance improves slightly to 28\%, suggesting that the previous drop might be due to context length limitations rather than semantic difficulty alone. These patterns validate our benchmark's difficulty levels while highlighting the importance of considering both semantic and context length effects.


\section{Conclusion}

In this work, we introduced SCALAR, a novel benchmark designed to evaluate LLMs long-context understanding while mitigating data contamination. By leveraging citations from scholarly papers, we generate high-quality ground truth labels while controlling task difficulty. Our experiments with state-of-the-art LLMs reveal that while models can effectively handle simple citation matching, they struggle with deeper comprehension of complex, context-rich references. SCALAR offers a sustainable and evolving benchmark to track advancements in long-context processing, providing insights for future model development.

\section*{Limitation}
SCALAR currently focuses on citation-based QA and cloze-style matching tasks, which capture a narrow but meaningful dimension of long-context understanding, identifying and reasoning over citation evidence in scholarly texts. It does not evaluate broader academic reasoning skills such as summarization, synthesis, or argument tracking.
SCALAR currently focuses on MCQA and cloze-style citation-matching tasks, which may not fully assess broader comprehension and reasoning abilities. Additionally, its scope is limited to computer science, restricting its applicability to other academic domains. To address these limitations, we plan to introduce diverse evaluation formats, while also expanding SCALAR to fields like biomedical and legal research for a more comprehensive assessment of long-context understanding.

\bibliography{anthology,custom}

@article{needle,
  title={{Needle In A Haystack} - Pressure Testing {LLM}s},
   author={Gregory Kamradt},
   year={2023},
   journal ={Github},  
   url={https://github.com/gkamradt/LLMTest_NeedleInAHaystack/tree/main},
}

@inproceedings{leval,
  title={L-eval: Instituting standardized evaluation for long context language models},
  author={An, Chenxin and Gong, Shansan and Zhong, Ming and Li, Mukai and Zhang, Jun and Kong, Lingpeng and Qiu, Xipeng},
  booktitle={ ICLR},
  year={2024}
}

@article{loogle,
  title={LooGLE: Can Long-Context Language Models Understand Long Contexts?},
  author={Li, Jiaqi and Wang, Mengmeng and Zheng, Zilong and Zhang, Muhan},
  journal={ arXiv:2311.04939},
  year={2023}
}

@article{inftybench,
      title={$\infty$Bench: Extending Long Context Evaluation Beyond 100K Tokens}, 
      author={Xinrong Zhang and Yingfa Chen and Shengding Hu and Zihang Xu and Junhao Chen and Moo Khai Hao and Xu Han and Zhen Leng Thai and Shuo Wang and Zhiyuan Liu and Maosong Sun},
      year={2024},
      journal={arXiv:2402.13718},
      archivePrefix={arXiv},
      primaryClass={cs.CL}
}

@article{bamboo,
  title={Bamboo: A comprehensive benchmark for evaluating long text modeling capacities of large language models},
  author={Dong, Zican and Tang, Tianyi and Li, Junyi and Zhao, Wayne Xin and Wen, Ji-Rong},
  journal={ arXiv:2309.13345},
  year={2023}
}

@misc{helmet,
      title={HELMET: How to Evaluate Long-Context Language Models Effectively and Thoroughly}, 
      author={Howard Yen and Tianyu Gao and Minmin Hou and Ke Ding and Daniel Fleischer and Peter Izsak and Moshe Wasserblat and Danqi Chen},
      year={2024},
      eprint={2410.02694},
      archivePrefix={arXiv},
      primaryClass={cs.CL},
      url={https://arxiv.org/abs/2410.02694}, 
}

@inproceedings{zeroscrolls,
    title = "{Z}ero{SCROLLS}: A Zero-Shot Benchmark for Long Text Understanding",
    author = "Shaham, Uri  and
      Ivgi, Maor  and
      Efrat, Avia  and
      Berant, Jonathan  and
      Levy, Omer",
    booktitle = " EMNLP ",
    year = "2023",
}

@article{longbench,
  title={{LongBench}: A Bilingual, Multitask Benchmark for Long Context Understanding},
  author={Bai, Yushi and others},
  journal={ arXiv:2308.14508},
  year={2023}
}

@article{babilong,
  title={BABILong: Testing the Limits of LLMs with Long Context Reasoning-in-a-Haystack},
  author={Kuratov, Yuri and Bulatov, Aydar and Anokhin, Petr and Rodkin, Ivan and Sorokin, Dmitry and Sorokin, Artyom and Burtsev, Mikhail},
  journal={arXiv preprint arXiv:2406.10149},
  year={2024}
}

@misc{ruler,
      title={RULER: What's the Real Context Size of Your Long-Context Language Models?}, 
      author={Cheng-Ping Hsieh and Simeng Sun and Samuel Kriman and Shantanu Acharya and Dima Rekesh and Fei Jia and Yang Zhang and Boris Ginsburg},
      year={2024},
      eprint={2404.06654},
      archivePrefix={arXiv},
      primaryClass={cs.CL},
      url={https://arxiv.org/abs/2404.06654}, 
}

@misc{litsearch,
      title={LitSearch: A Retrieval Benchmark for Scientific Literature Search}, 
      author={Anirudh Ajith and Mengzhou Xia and Alexis Chevalier and Tanya Goyal and Danqi Chen and Tianyu Gao},
      year={2024},
      eprint={2407.18940},
      archivePrefix={arXiv},
      primaryClass={cs.IR},
      url={https://arxiv.org/abs/2407.18940}, 
}

@misc{citebench,
      title={CiteBench: A benchmark for Scientific Citation Text Generation}, 
      author={Martin Funkquist and Ilia Kuznetsov and Yufang Hou and Iryna Gurevych},
      year={2023},
      eprint={2212.09577},
      archivePrefix={arXiv},
      primaryClass={cs.CL},
      url={https://arxiv.org/abs/2212.09577}, 
}

@article{achiam2023gpt,
  title={Gpt-4 technical report},
  author={Achiam, Josh and Adler, Steven and Agarwal, Sandhini and Ahmad, Lama and Akkaya, Ilge and Aleman, Florencia Leoni and Almeida, Diogo and Altenschmidt, Janko and Altman, Sam and Anadkat, Shyamal and others},
  journal={arXiv preprint arXiv:2303.08774},
  year={2023}
}

@article{yang2025qwen2,
  title={Qwen2. 5-1M Technical Report},
  author={Yang, An and Yu, Bowen and Li, Chengyuan and Liu, Dayiheng and Huang, Fei and Huang, Haoyan and Jiang, Jiandong and Tu, Jianhong and Zhang, Jianwei and Zhou, Jingren and others},
  journal={arXiv preprint arXiv:2501.15383},
  year={2025}
}

@misc{anthropic2024,
  author = {Anthropic},
  title = {Introducing the next generation of Claude},
  year = {2024},
  url = {https://www.anthropic.com/news/claude-3-family}
}

@article{dubey2024llama,
  title={The llama 3 herd of models},
  author={Dubey, Abhimanyu and Jauhri, Abhinav and Pandey, Abhinav and Kadian, Abhishek and Al-Dahle, Ahmad and Letman, Aiesha and Mathur, Akhil and Schelten, Alan and Yang, Amy and Fan, Angela and others},
  journal={arXiv preprint arXiv:2407.21783},
  year={2024}
}

@misc{s2orc,
      title={S2ORC: The Semantic Scholar Open Research Corpus}, 
      author={Kyle Lo and Lucy Lu Wang and Mark Neumann and Rodney Kinney and Dan S. Weld},
      year={2020},
      eprint={1911.02782},
      archivePrefix={arXiv},
      primaryClass={cs.CL},
      url={https://arxiv.org/abs/1911.02782}, 
}

@misc{pile,
      title={The Pile: An 800GB Dataset of Diverse Text for Language Modeling}, 
      author={Leo Gao and Stella Biderman and Sid Black and Laurence Golding and Travis Hoppe and Charles Foster and Jason Phang and Horace He and Anish Thite and Noa Nabeshima and Shawn Presser and Connor Leahy},
      year={2020},
      eprint={2101.00027},
      archivePrefix={arXiv},
      primaryClass={cs.CL},
      url={https://arxiv.org/abs/2101.00027}, 
}

@InProceedings{10.1007/978-3-030-99736-6_19,
author="Gu, Nianlong
and Gao, Yingqiang
and Hahnloser, Richard H. R.",
editor="Hagen, Matthias
and Verberne, Suzan
and Macdonald, Craig
and Seifert, Christin
and Balog, Krisztian
and N{\o}rv{\aa}g, Kjetil
and Setty, Vinay",
title="Local Citation Recommendation with Hierarchical-Attention Text Encoder and SciBERT-Based Reranking",
booktitle="Advances in Information Retrieval",
year="2022",
publisher="Springer International Publishing",
address="Cham",
pages="274--288",
isbn="978-3-030-99736-6"
}

@article{F_rber_2020,
   title={Citation recommendation: approaches and datasets},
   volume={21},
   ISSN={1432-1300},
   url={http://dx.doi.org/10.1007/s00799-020-00288-2},
   DOI={10.1007/s00799-020-00288-2},
   number={4},
   journal={International Journal on Digital Libraries},
   publisher={Springer Science and Business Media LLC},
   author={Färber, Michael and Jatowt, Adam},
   year={2020},
   month=aug, pages={375–405} }

@misc{abdin2024phi4technicalreport,
      title={Phi-4 Technical Report}, 
      author={Marah Abdin and Jyoti Aneja and Harkirat Behl and Sébastien Bubeck and Ronen Eldan and Suriya Gunasekar and Michael Harrison and Russell J. Hewett and Mojan Javaheripi and Piero Kauffmann and James R. Lee and Yin Tat Lee and Yuanzhi Li and Weishung Liu and Caio C. T. Mendes and Anh Nguyen and Eric Price and Gustavo de Rosa and Olli Saarikivi and Adil Salim and Shital Shah and Xin Wang and Rachel Ward and Yue Wu and Dingli Yu and Cyril Zhang and Yi Zhang},
      year={2024},
      eprint={2412.08905},
      archivePrefix={arXiv},
      primaryClass={cs.CL},
      url={https://arxiv.org/abs/2412.08905}, 
}

@techreport{claude21modelcard, title = {Claude 2.1 Model Card}, author = {Claude}, year = {2023}, institution = {Claude Inc.}, url = {https://claude.ai/model-card/claude-2-1} }

@article{team2024gemini,
  title={Gemini 1.5: Unlocking multimodal understanding across millions of tokens of context},
  author={Team, Gemini and Georgiev, Petko and Lei, Ving Ian and Burnell, Ryan and Bai, Libin and Gulati, Anmol and Tanzer, Garrett and Vincent, Damien and Pan, Zhufeng and Wang, Shibo and others},
  journal={arXiv preprint arXiv:2403.05530},
  year={2024}
}

@misc{longbenchv2,
      title={LongBench v2: Towards Deeper Understanding and Reasoning on Realistic Long-context Multitasks}, 
      author={Yushi Bai and Shangqing Tu and Jiajie Zhang and Hao Peng and Xiaozhi Wang and Xin Lv and Shulin Cao and Jiazheng Xu and Lei Hou and Yuxiao Dong and Jie Tang and Juanzi Li},
      year={2025},
      eprint={2412.15204},
      archivePrefix={arXiv},
      primaryClass={cs.CL},
      url={https://arxiv.org/abs/2412.15204}, 
}

@inproceedings{qasper,
    title = "A Dataset of Information-Seeking Questions and Answers Anchored in Research Papers",
    author = "Dasigi, Pradeep  and
      Lo, Kyle  and
      Beltagy, Iz  and
      Cohan, Arman  and
      Smith, Noah A.  and
      Gardner, Matt",
    editor = "Toutanova, Kristina  and
      Rumshisky, Anna  and
      Zettlemoyer, Luke  and
      Hakkani-Tur, Dilek  and
      Beltagy, Iz  and
      Bethard, Steven  and
      Cotterell, Ryan  and
      Chakraborty, Tanmoy  and
      Zhou, Yichao",
    booktitle = "Proceedings of the 2021 Conference of the North American Chapter of the Association for Computational Linguistics: Human Language Technologies",
    month = jun,
    year = "2021",
    address = "Online",
    publisher = "Association for Computational Linguistics",
    url = "https://aclanthology.org/2021.naacl-main.365/",
    doi = "10.18653/v1/2021.naacl-main.365",
    pages = "4599--4610"
}

@article{Lala2023PaperQA,
  title     = {PaperQA: Retrieval-Augmented Generative Agent for Scientific Research},
  author    = {Jakub L\'ala and Odhran O’Donoghue and Aleksandar Shtedritski and Sam Cox and Samuel G Rodriques and Andrew D White},
  year      = {2023},
}

@article{touvron2023llama2,
  title={Llama 2: Open foundation and fine-tuned chat models},
  author={Touvron, Hugo and Martin, Louis and Stone, Kevin and Albert, Peter and Almahairi, Amjad and Babaei, Yasmine and Bashlykov, Nikolay and Batra, Soumya and Bhargava, Prajjwal and Bhosale, Shruti and others},
  journal={arXiv preprint arXiv:2307.09288},
  year={2023}
}

@article{guo2025deepseek,
  title={Deepseek-r1: Incentivizing reasoning capability in llms via reinforcement learning},
  author={Guo, Daya and Yang, Dejian and Zhang, Haowei and Song, Junxiao and Zhang, Ruoyu and Xu, Runxin and Zhu, Qihao and Ma, Shirong and Wang, Peiyi and Bi, Xiao and others},
  journal={arXiv preprint arXiv:2501.12948},
  year={2025}
}

@misc{kuratov2024searchneedles11mhaystack,
      title={In Search of Needles in a 11M Haystack: Recurrent Memory Finds What LLMs Miss}, 
      author={Yuri Kuratov and Aydar Bulatov and Petr Anokhin and Dmitry Sorokin and Artyom Sorokin and Mikhail Burtsev},
      year={2024},
      eprint={2402.10790},
      archivePrefix={arXiv},
      primaryClass={cs.CL},
      url={https://arxiv.org/abs/2402.10790}, 
}

@inproceedings{
wang2024needle,
title={Needle In A Multimodal Haystack},
author={Weiyun Wang and Shuibo Zhang and Yiming Ren and Yuchen Duan and Tiantong Li and Shuo Liu and Mengkang Hu and Zhe Chen and Kaipeng Zhang and Lewei Lu and Xizhou Zhu and Ping Luo and Yu Qiao and Jifeng Dai and Wenqi Shao and Wenhai Wang},
booktitle={The Thirty-eight Conference on Neural Information Processing Systems Datasets and Benchmarks Track},
year={2024},
url={https://openreview.net/forum?id=U2pNwSuQqD}
}

@misc{roberts2024needlethreadingllmsfollow,
      title={Needle Threading: Can LLMs Follow Threads through Near-Million-Scale Haystacks?}, 
      author={Jonathan Roberts and Kai Han and Samuel Albanie},
      year={2024},
      eprint={2411.05000},
      archivePrefix={arXiv},
      primaryClass={cs.CL},
      url={https://arxiv.org/abs/2411.05000}, 
}

@inproceedings{
chang2024booookscore,
title={BooookScore: A systematic exploration of book-length summarization in the era of {LLM}s},
author={Yapei Chang and Kyle Lo and Tanya Goyal and Mohit Iyyer},
booktitle={The Twelfth International Conference on Learning Representations},
year={2024},
url={https://openreview.net/forum?id=7Ttk3RzDeu}
}

@inproceedings{reddy-etal-2024-docfinqa,
    title = "{D}oc{F}in{QA}: A Long-Context Financial Reasoning Dataset",
    author = "Reddy, Varshini  and
      Koncel-Kedziorski, Rik  and
      Lai, Viet Dac  and
      Krumdick, Michael  and
      Lovering, Charles  and
      Tanner, Chris",
    editor = "Ku, Lun-Wei  and
      Martins, Andre  and
      Srikumar, Vivek",
    booktitle = "Proceedings of the 62nd Annual Meeting of the Association for Computational Linguistics (Volume 2: Short Papers)",
    month = aug,
    year = "2024",
    address = "Bangkok, Thailand",
    publisher = "Association for Computational Linguistics",
    url = "https://aclanthology.org/2024.acl-short.42/",
    doi = "10.18653/v1/2024.acl-short.42",
    pages = "445--458",
}

@misc{wang2024mathhayautomatedbenchmarklongcontext,
      title={MathHay: An Automated Benchmark for Long-Context Mathematical Reasoning in LLMs}, 
      author={Lei Wang and Shan Dong and Yuhui Xu and Hanze Dong and Yalu Wang and Amrita Saha and Ee-Peng Lim and Caiming Xiong and Doyen Sahoo},
      year={2024},
      eprint={2410.04698},
      archivePrefix={arXiv},
      primaryClass={cs.CL},
      url={https://arxiv.org/abs/2410.04698}, 
}

@misc{zhao2024docmathevalevaluatingmathreasoning,
      title={DocMath-Eval: Evaluating Math Reasoning Capabilities of LLMs in Understanding Long and Specialized Documents}, 
      author={Yilun Zhao and Yitao Long and Hongjun Liu and Ryo Kamoi and Linyong Nan and Lyuhao Chen and Yixin Liu and Xiangru Tang and Rui Zhang and Arman Cohan},
      year={2024},
      eprint={2311.09805},
      archivePrefix={arXiv},
      primaryClass={cs.CL},
      url={https://arxiv.org/abs/2311.09805}, 
}

@misc{laban2024summaryhaystackchallengelongcontext,
      title={Summary of a Haystack: A Challenge to Long-Context LLMs and RAG Systems}, 
      author={Philippe Laban and Alexander R. Fabbri and Caiming Xiong and Chien-Sheng Wu},
      year={2024},
      eprint={2407.01370},
      archivePrefix={arXiv},
      primaryClass={cs.CL},
      url={https://arxiv.org/abs/2407.01370}, 
}

@misc{wang2024leavedocumentbehindbenchmarking,
      title={Leave No Document Behind: Benchmarking Long-Context LLMs with Extended Multi-Doc QA}, 
      author={Minzheng Wang and Longze Chen and Cheng Fu and Shengyi Liao and Xinghua Zhang and Bingli Wu and Haiyang Yu and Nan Xu and Lei Zhang and Run Luo and Yunshui Li and Min Yang and Fei Huang and Yongbin Li},
      year={2024},
      eprint={2406.17419},
      archivePrefix={arXiv},
      primaryClass={cs.CL},
      url={https://arxiv.org/abs/2406.17419}, 
}

@misc{qiu2024clongevalchinesebenchmarkevaluating,
      title={CLongEval: A Chinese Benchmark for Evaluating Long-Context Large Language Models}, 
      author={Zexuan Qiu and Jingjing Li and Shijue Huang and Xiaoqi Jiao and Wanjun Zhong and Irwin King},
      year={2024},
      eprint={2403.03514},
      archivePrefix={arXiv},
      primaryClass={cs.CL},
      url={https://arxiv.org/abs/2403.03514}, 
}

@misc{zhang2024marathonracerealmlong,
      title={Marathon: A Race Through the Realm of Long Context with Large Language Models}, 
      author={Lei Zhang and Yunshui Li and Ziqiang Liu and Jiaxi yang and Junhao Liu and Longze Chen and Run Luo and Min Yang},
      year={2024},
      eprint={2312.09542},
      archivePrefix={arXiv},
      primaryClass={cs.CL},
      url={https://arxiv.org/abs/2312.09542}, 
}

@misc{song2024countingstarsmultievidencepositionawarescalable,
      title={Counting-Stars: A Multi-evidence, Position-aware, and Scalable Benchmark for Evaluating Long-Context Large Language Models}, 
      author={Mingyang Song and Mao Zheng and Xuan Luo},
      year={2024},
      eprint={2403.11802},
      archivePrefix={arXiv},
      primaryClass={cs.CL},
      url={https://arxiv.org/abs/2403.11802}, 
}

@misc{openr1,
    title = {Open R1: A fully open reproduction of DeepSeek-R1},
    url = {https://github.com/huggingface/open-r1},
    author = {Hugging Face},
    month = {January},
    year = {2025}
}

@misc{yu2025dapoopensourcellmreinforcement,
      title={DAPO: An Open-Source LLM Reinforcement Learning System at Scale}, 
      author={Qiying Yu and Zheng Zhang and Ruofei Zhu and Yufeng Yuan and Xiaochen Zuo and Yu Yue and Tiantian Fan and Gaohong Liu and Lingjun Liu and Xin Liu and Haibin Lin and Zhiqi Lin and Bole Ma and Guangming Sheng and Yuxuan Tong and Chi Zhang and Mofan Zhang and Wang Zhang and Hang Zhu and Jinhua Zhu and Jiaze Chen and Jiangjie Chen and Chengyi Wang and Hongli Yu and Weinan Dai and Yuxuan Song and Xiangpeng Wei and Hao Zhou and Jingjing Liu and Wei-Ying Ma and Ya-Qin Zhang and Lin Yan and Mu Qiao and Yonghui Wu and Mingxuan Wang},
      year={2025},
      eprint={2503.14476},
      archivePrefix={arXiv},
      primaryClass={cs.LG},
      url={https://arxiv.org/abs/2503.14476}, 
}

@misc{openai2023tiktoken,
  author       = {OpenAI},
  title        = {tiktoken},
  howpublished = {\url{https://github.com/openai/tiktoken}},
  year         = {2023},
}
\bibliographystyle{acl_natbib}

\clearpage
\newpage
\onecolumn
\appendix

\section{Difficulty Control in SCALAR}\label{sec:difficulty_config}
\begin{figure*}[h]    
\centering
    \includegraphics[width=1.0\linewidth]{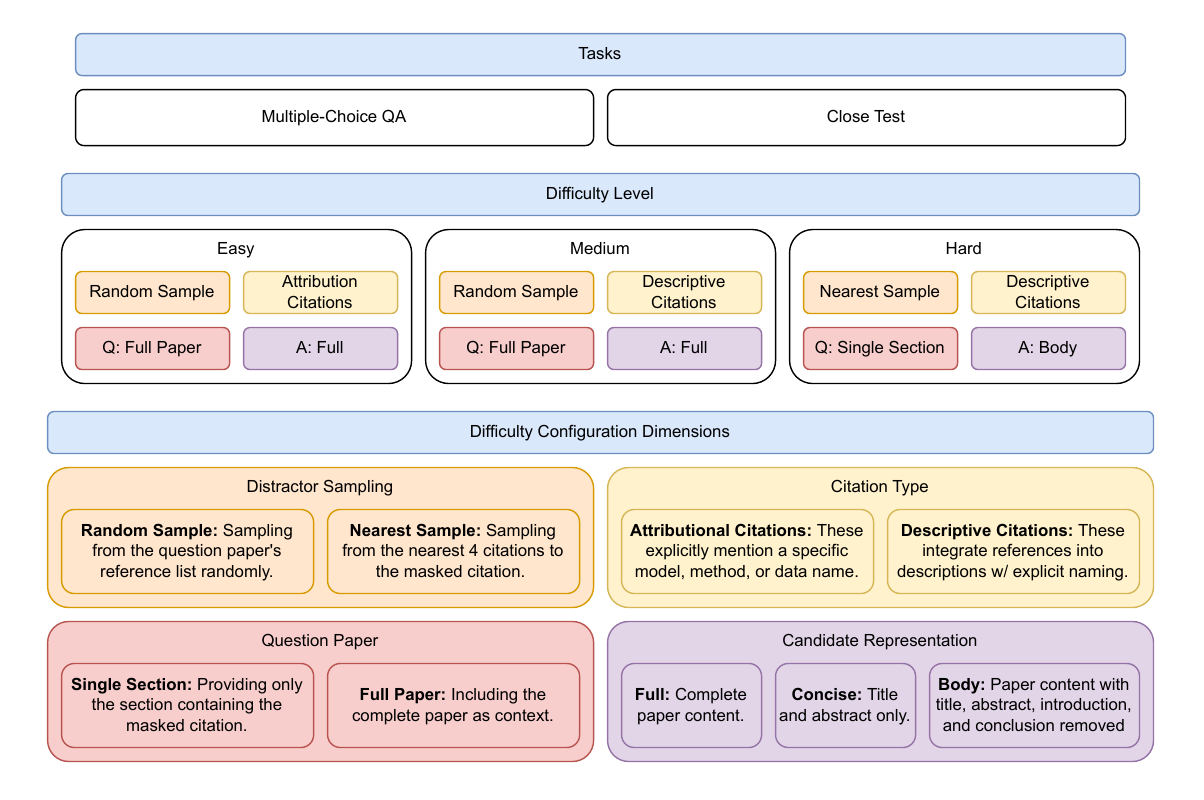}
    \caption{Overview of our configurable framework for difficulty control. The benchmark supports two tasks (Multiple-Choice QA and Cloze Test) and categorizes difficulty into three levels by varying distractor sampling strategies, citation types, question paper context, and candidate representations. These dimensions allow for fine-grained semantic and contextual adjustments, as illustrated in the diagram.}
	\label{fig:difficulty_config}
\end{figure*}

\section{Model Context Length and Price}\label{sec:context_price}
We list the context length and price of the models in this paper as in~\tabref{tab:context_price}.

\begin{table}[h]
\centering
\begin{tabular}{lcl}
\toprule
Model & Context & Price \\
\midrule
Llama-3.1-8B            & 128K  & \$0.05* \\
Llama-3.1-70B           & 128K  & \$0.65* \\
Llama-3.3-70B           & 128K  & \$0.65* \\
Qwen2.5-7B-1M           & 1M    & \$0.05* \\
QwQ-32B                 & 128K  & \$0.1* \\
Qwen3-8B                & 128K  & \$0.035* \\
Phi-3-Mini-128K-Instruct   & 128K & \$0.03* \\
Phi-3-Medium-128K-Instruct & 128K & \$0.10* \\
Claude-3.5-Haiku        & 200K  & \$0.80 \\
GPT-4o-Mini             & 128K  & \$0.15 \\
GPT-4o                  & 128K  & \$2.50 \\
\bottomrule
\end{tabular}
\caption{Model context length and price per 1M tokens. * denotes price estimated via third-party inference service.}
\label{tab:context_price}
\end{table}

\section{Prompt use in SCALAR}
\label{sec:prompt}

The prompt template used for the multiple choice QA and cloze test are in \Cref{fig:prompt_mc} and \Cref{fig:prompt_cloze}, respectively.

\begin{figure*}[h]
	\small
	\begin{tcolorbox}[colframe=white, left=2.5mm, right=1.5mm]
You are given a paper with a placeholder ``**[MASKED\_CITATION]**'' in
its content. Your task is to select the most appropriate reference
from the provided reference list to replace the mask.\\
- The paper content is enclosed within <Paper> and </Paper>.\\
- The reference list is enclosed within <References> and </References>,\\
with each reference wrapped in <Candidate> and </Candidate>.\\
- After selecting the best-suited reference, output the index of that
reference in the following format:\\
<answer>index</answer>.\\
<Paper>\\
\blue{... BERT (*[MASKED\_CITATION]**) or ...} \\
</Paper>\\
<References>\\
<Candidate>Candidate [0]:\\
\blue{... candidate content ...}\\
</Candidate>\\
<Candidate>Candidate [1]:\\
\blue{... candidate content ...}\\
</Candidate>\\
<Candidate>Candidate [2]:\\
\blue{... candidate content ...}\\
</Candidate>\\
<Candidate>Candidate [3]:\\
\blue{... candidate content ...}\\
</Candidate>\\
</References>\\
Remember to output the index of the selected reference enclosed within
<answer> and </answer>.\\
	\end{tcolorbox}
	\caption{The prompt template used for the multiple choice QA.}
	\label{fig:prompt_mc}
\end{figure*}

\begin{figure*}[h]
	\small
	\begin{tcolorbox}[colframe=white, left=2.5mm, right=1.5mm]
You are given a paper with four placeholders **[MASKED\_CITATION\_0]**, **[MASKED\_CITATION\_1]**, **[MASKED\_CITATION\_2]**, and **[MASKED\_CITATION\_3]**, each hiding a citation, plus exactly four reference candidates. \\
- The paper content is enclosed within <Paper> and </Paper>. \\
- The reference list is enclosed within <References> and </References>, with each reference wrapped in <Candidate> and </Candidate>. \\
- Return **one** <answer> tag with four separated integers giving the candidate index (0-3) for placeholders in the order of **[MASKED\_CITATION\_0]**, **[MASKED\_CITATION\_1]**, **[MASKED\_CITATION\_2]**, and **[MASKED\_CITATION\_3]**. For example: <answer> \\
2 \\
1 \\
3 \\
0 \\
</answer>. \\
<Paper>\\
\blue{... BERT (*[MASKED\_CITATION\_0]**) or ...} \\
</Paper>\\
<References>\\
<Candidate>Candidate [0]:\\
\blue{... candidate content ...}\\
</Candidate>\\
<Candidate>Candidate [1]:\\
\blue{... candidate content ...}\\
</Candidate>\\
<Candidate>Candidate [2]:\\
\blue{... candidate content ...}\\
</Candidate>\\
<Candidate>Candidate [3]:\\
\blue{... candidate content ...}\\
</Candidate>\\
</References>\\
Remember: output four integers wrapped inside a single <answer> tag.\\
	\end{tcolorbox}
	\caption{The prompt template used for Cloze-style test.}
	\label{fig:prompt_cloze}
\end{figure*}

\end{document}